\documentclass[sigconf, nonacm]{acmart}

\usepackage{amsmath}
\usepackage{tikz}
\usepackage[ruled,vlined,linesnumbered]{algorithm2e}

\usepackage{fixltx2e}

\usepackage{filecontents}
\usepackage{kotex}
\usepackage{comment}
\usepackage{enumitem}
\usepackage{color}
\usepackage{booktabs}
\usepackage{arydshln}
\usepackage[normalem]{ulem}

\newcommand{\eat}[1]{}

\newcommand{\revsched}{reverse first-{\it k} scheduling}

\begin{document}

\date{} 

\title{Scheduling Optimization Techniques for Neural Network Training}

\author{Hyungjun Oh}
\affiliation{%
  \institution{\small Hanyang University}
  \country{Korea}
}
\email{tcpip15@hanyang.ac.kr}

\author{Junyeol Lee}
\affiliation{%
    \institution{\small Hanyang University}
    \country{Korea}
}
\email{shie007@hanyang.ac.kr}

\author{HyeongJu Kim}
\affiliation{%
    \institution{\small Hanyang University}
    \country{Korea}
}
\email{gudwn0520@hanyang.ac.kr}

\author{Jiwon Seo}
\authornote{Corresponding author and principal investigator}
\affiliation{%
    \institution{\small Hanyang University}
    \country{Korea}
}
\email{seojiwon@hanyang.ac.kr}

\begin{abstract}
Neural network training requires a large amount of computation
and thus GPUs are often used for the acceleration.
While they improve the performance, GPUs are underutilized during the training.
This paper proposes {\it out-of-order (ooo) backprop}, an effective
scheduling technique for neural network training.
By exploiting the dependencies of gradient computations,
ooo backprop enables to reorder their executions to make the most of the GPU resources.
We show that the GPU utilization in single-GPU, data-parallel, and pipeline-parallel training
can be commonly improve by applying ooo backprop and prioritizing critical operations.
We propose three scheduling algorithms based on ooo backprop.
For single-GPU training, we schedule with multi-stream out-of-order computation to mask
the kernel launch overhead. In data-parallel training, we reorder the gradient computations to
maximize the overlapping of computation and parameter communication; 
in pipeline-parallel training, we prioritize critical gradient computations
to reduce the pipeline stalls.
We evaluate our optimizations with twelve neural networks
including a light-weight computer vision model\,(MobileNet) and large NLP models\,(BERT and GPT-3) with up to forty eight V100 GPUs.
Our scheduling algorithms effectively improve the performance of
single-GPU training as well as data- and pipeline-parallel training.
Compared to the respective state of the art training systems,
the throughput is substantially improved for single-GPU, data-parallel,
and pipeline-parallel training.

\end{abstract}

\maketitle

\settopmatter{printacmref=false, printccs=false, printfolios=false}

\section{Introduction}

Deep neural networks (DNNs) are now widely used in many domains.
Because training and running neural networks are computationally expensive,
GPUs are commonly used for the acceleration.
While they substantially speedup the performance,
GPUs are often underutilized when running neural network tasks.
At a single GPU level, many of neural network kernels
have low GPU resource utilization\,\cite{song2016bridging, li2016performance,zhou2017performance};
at a cluster level, half of the GPUs running neural network tasks are 
idle, wasting their computation cycles\,\cite{gpucluster2019}.

Hardware resource utilization is not a new problem.
In the 80s and 90s, the increasing transistor density and
clock frequency of CPUs made it challenging to efficiently utilize
CPU resources.
To improve the utilization efficiency, instruction-level parallelism (ILP)
had been extensively studied in both hardware and software aspects.
Techniques such as instruction pipelining and out-of-order execution are proposed and applied to increase the degree of ILP and thus improve the utilization of 
functional units in CPUs.

Inspired by the past studies on ILP and carefully investigating DNN tasks,
we propose scheduling optimizations for neural network training.
Although the GPU underutilization for single- and multi-GPU DNN training is
caused by different reasons, we found out that their performance can be significantly improved 
by scheduling their operations efficiently.
We proposed three scheduling algorithms for single- and multi-GPU (data- and pipeline-parallel) training.
The three algorithms, while they differ largely in details, apply the same principle of
prioritizing critical operations and increasing execution concurrency
to improve the GPU utilization and the training performance.

All our scheduling algorithms for single and multi-GPU training is based on 
our novel scheduling technique, which we call {\it out-of-order backprop}.
Although existing deep learning systems perform backpropagation 
strictly in the reverse order of the network layouts, 
we observed that a subset of gradient computations may be executed in an out-of-order manner.
By exploiting this property, we schedule the gradient computations
such that the critical ones are executed with higher priorities.
For single-GPU training, 
out-of-order backprop helps to mask the kernel launch overhead;
for data-parallel and pipeline-parallel training, the technique helps to
maximize the overlapping of inter-GPU communication with gradient computations.

This paper contributes out-of-order backprop and the scheduling algorithms based on this technique.
We summarize our specific contributions in the following.

\noindent
{\bf Concept of out-of-order backprop.}
We propose {\it out-of-order backprop} as a general principle
for scheduling the computations in DNN training.
Exploiting the computation dependencies in the training, 
out-of-order backprop enables the execution of the gradient computations 
out of their layout order so that more critical computations
are executed with higher priorities.
\eat{Like out-of-order execution in CPU microarchitecture,
out-of-order backprop helps to improve GPU resource utilization.}

\noindent
{\bf Scheduling algorithms for single and multi-GPU training.}
We designed three novel scheduling algorithms 
based on out-of-order backprop and the list scheduling technique.
For single-GPU training, we schedule weight- and output-gradient computations
in multiple streams and in an out-of-order manner.
For data-parallel training, our scheduling algorithm reorders the gradient computations to maximize
the overlapping of the communication and computation.
For pipeline-parallel training, we prioritize critical gradient computations
to reduce the pipeline stalls.
All our scheduling algorithms make efficient use of gradient computation reordering (i.e., out-of-order backprop), which no prior work had exploited for both single- or multi-GPU training.

\noindent
{\bf Implementation in real-world deep learning systems.}
We implement out-of-order backprop and our scheduling algorithms in 
TensorFlow, a widely-used deep learning system.
We modified TensorFlow's execution engine and its compiler XLA
to implement our scheduling techniques; 
we also added an efficient support for an auxiliary GPU stream 
to concurrently execute a subset of gradient computations.
Moreover, we implemented our scheduling algorithms in BytePS,
the state of the art parameter-communication system for distributed neural network training\,\cite{byteps}.

\noindent
{\bf Extensive evaluation and availability.}
We evaluate out-of-order backprop and the scheduling algorithms with twelve neural network models in computer vision and natural language processing.
The evaluation is performed on three different GPU models, with four types of network interconnect, and on a cluster of up to forty eight GPUs.
For all (single- and multi-GPU) training methods, our scheduling algorithms largely improve the performance, compared to the respective state of the art systems.
For single-GPU training, our scheduling algorithm using multi-stream out-of-order computation improves the training performance by 1.03--1.58$\times$ over TensorFlow XLA; 
compared to Nimble, a state of the art deep learning execution engine, 
we exceed its performance by 1.28$\times$ on average.
For data-parallel training, our technique that prioritizes critical computations 
outperforms BytePS by 1.10--1.27$\times$ on a cluster with sixteen to forty eight GPUs.
For pipeline-parallel training, we outperform GPipe by 1.41--1.99$\times$ on a cluster with 
four to thirty six GPUs;
compared to PipeDream that applies {\it weight stashing} and thus changes the semantics of the training,
our execution runs 1.31$\times$ faster on average. 
We report the analysis of our performance improvements for the three training methods.
We plan to open-source our implementations and scheduling algorithms in TensorFlow and BytePS as well as all the execution schedules for the evaluated neural network models.

\begin{figure}[t]
\centering
\includegraphics[width=3.2in]{img/kernel_issue_overhead.png}
\vspace{-2mm}
\caption{Kernel issue overhead of deep learning systems for convolution and ReLU operations in DenseNet-121.}
\label{fig:exec-overhead}
\vspace{-2mm}
\end{figure}

The rest of the paper is organized as follows.
Section~\ref{sec:underutil_gpu} reviews the existing GPU underutilization problems.
Section~\ref{sec:ooobp} presents out-of-order backprop, our core scheduling technique.
Then Section~\ref{sec:single-gpu} and \ref{sec:multi-gpu} describe
our scheduling algorithms for single-GPU, data-parallel, and pipeline-parallel training.
Section~\ref{sec:impl} explains our implementation in TensorFlow.
Section~\ref{sec:eval} evaluates out-of-order backprop and our scheduling algorithms based on the technique.
Section~\ref{sec:related} discusses related work and 
Section~\ref{sec:conclude} concludes.

\section{GPU Underutilization Problems}
\label{sec:underutil_gpu}

DNN training requires a large amount of computation
and GPUs are widely used for the acceleration.
Although they largely speed up the training, 
GPUs are often underutilized during the training\,\cite{song2016bridging, li2016performance,zhou2017performance, gpucluster2019}. 
Here we review the GPU underutilization problems in single- and multi-GPU training;
then we describe how these problems can be commonly formulated as a common scheduling optimization problem.

\begin{figure}[tb]
\centering
\includegraphics[width=3.2in]{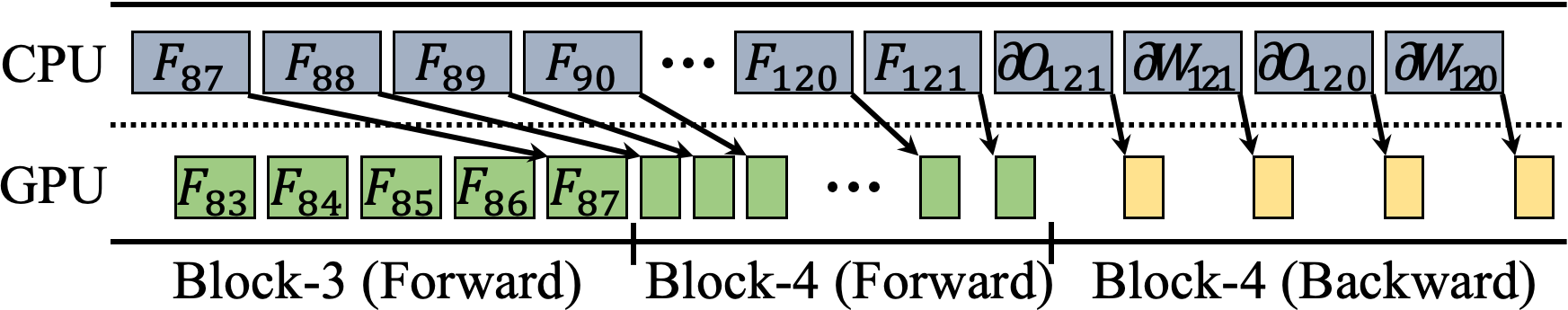}
\vspace{-1mm}
\caption{Execution timeline for DenseNet-121 (Intel Xeon E5-2698, NVIDIA V100). The kernel issue overhead (top) and their actual executions on GPU\,(bottom) are shown.
\label{fig:dense121_execution_timeline}} 
\end{figure}

\vspace{1mm}
\noindent
{\bf Analysis for single-GPU training.}
In single-GPU training, the GPU underutilization is mainly caused by
kernel issue/execution overhead and idling SMs (Stream Multiprocessors) during kernel executions.
Let us first consider the kernel issue overhead.
In deep learning systems such as TensorFlow,
DNN training is represented by (implicit or explicit) computation graphs
with DNN operations and their dependencies.
The systems have an {\it executor} that traverses the graph and asynchronously issues the GPU kernels.
If the executor's latency of issuing the kernels is longer than
their executions on GPU, this overhead may become the performance bottleneck.

In our preliminary experiments, we measured the kernel issue overhead 
in three systems (TensorFlow, PyTorch, and MXNet).
For many DNN models, the overhead is the bottleneck
of the training. Particularly, recent convolutional neural networks (CNNs), 
such as DenseNet or MobileNet,
are largely affected by this overhead as they have many light-weight convolutions.
Figure~\ref{fig:exec-overhead} shows the kernel issue overhead for 
DenseNet-121\,\cite{huang2017densely} (on Intel Xeon E5-2698 and NVIDIA V100).
For the convolutions in DenseBlock-3 and 4, their issue overhead is up to 4$\times$ larger than their execution times; since the two DenseBlocks take up two thirds of the total execution, this overhead is critical.

Specifically, Figure~\ref{fig:dense121_execution_timeline} shows the (simplified) actual timeline for training DenseNet-121 in TensorFlow;
part of the forward and backward computation is shown. 
For DenseBlock-3's forward computation, the kernel issue overhead is masked by the previously issued kernels. However, the masking effect disappears by the end of the next DenseBlock, 
when the idle time between the kernels increases.
This overhead is also recently reported in other studies\,\cite{kwon2020nimble, narayanan2018accelerating}.

Moreover, we also observed GPU's kernel execution overhead. 
Even if the kernel issue overhead is completely masked, there is
1--2$\mu s$ idle time between the kernel executions (e.g. forward computation of DenseBlock-3 in Figure~\ref{fig:dense121_execution_timeline}). 
This overhead is caused by the GPU's execution engine for setting up the SMs
for the kernel execution\,\cite{tanasic2014enabling, wong2010demystifying}.
For short-running kernels the overhead is non-trivial; e.g., for the convolution kernels in DenseBlock-3 and 4, 
their execution times are 15--40$\mu s$ and thus the kernel execution overhead is 3--13\% of their execution.

The other cause of the GPU underutilization is idling SMs during and end of kernel executions.
During the execution of a kernel, its thread blocks are scheduled and executed in the SMs.
If the kernel has a smaller number of thread blocks than the number that the SMs are capable of running at once, the SMs may not be fully utilized.
For example, for the weight gradient kernels in DenseBlock-4 (on V100 GPU),
half of the kernels run with the same configuration of 448 thread blocks. However, the SMs
are capable of running 1,520 of the thread blocks and thus they are underutilized
in terms of the thread block capacity. 
Also, consider the end of the kernel execution when its last thread blocks are scheduled.
At this point, the SMs are likely to be underutilized 
because they are running fewer number of thread blocks than their capacity;
this problem is also referred to as tail effect or tail underutilization\,\cite{nvidia_tail_effect,gtc12_tail_effect}.

\vspace{2mm}
\noindent
{\bf Analysis for GPU clusters.}
It is reported that the utilization of GPU clusters is less than 52\% for neural network tasks\,\cite{gpucluster2019}.
The low cluster utilization is caused by many factors, such as
inefficiency of task scheduling or interference of tasks.
The primary reason, however, 
is the communication overhead in data-parallel training and
the pipeline stalls in model-parallel (or pipeline-parallel) training.

In data-parallel training, workers communicate to synchronize their weight parameters in each training iteration. 
Because the forward computation of the next iteration blocks until the parameter synchronization 
is completed, GPU cycles are wasted during the synchronization.
It is reported that the wasted GPU cycles from the synchronization is 15--30\% of the total execution time
if wait-free backpropagation is applied to overlap the synchronization and gradient computation\,\cite{poseidon}. 
Prioritization of parameter communication in critical path\,\cite{jayarajan2019priority, hashemi2018tictac,byteps,peng2019generic,li13pytorch} may further reduce the cost but 
still the overhead is 10--25\% as our evaluation in Section~\ref{sec:data-parallel} shows.
Recently, asynchronous parameter communication schemes are proposed to improve
the training throughput\,\cite{adpsgd2018, ssp2013, easgd2015}.
However, because they may incur accuracy loss, those asynchronous schemes are less widely applied in practice\,\cite{wongpanich2020rethinking,gupta2016model,Ko2021An}.

Now let us consider model-parallel training.
For large models such as GPT, cross-layer model-parallelism is 
commonly used, where each layer is assigned to one of working GPUs for training,
which performs the layer's forward and backward computation.
Due to the computation dependency,
only a subset of the GPUs perform computation at once.
That is, in both forward and backward propagation, most of the GPUs are 
stalled waiting for the computation result from the GPUs assigned with the 
earlier (or later, in backpropagation) layers.

To mitigate the cost of the execution stalls, pipelining of the GPU computations
is further proposed\,\cite{gpipe19, pipedream19}.
In pipeline-parallel training, input data (i.e. mini-batch) is split into
{\it micro-batches}, which are sequentially fed to the GPUs 
for the concurrent executions of the GPUs.
Although using micro-batches increases the number of concurrently in-use GPUs,
still a substantial portion of the GPUs are idle during the training 
as we show in Section~\ref{sec:eval-pipepar}.
Recently PipeDream proposed to train with multiple versions of weight parameters 
to increase the overlapping the computations of different micro-batches\,\cite{gpipe19, pipedream19}.
However, this technique causes parameter staleness in a similar manner
to the asynchronous communication schemes in data-parallel training,
and thus it may negatively affect the learning efficiency\,\cite{ho2013more, dai2018toward}.

\vspace{2mm}
\noindent
{\bf Formulation of a common optimization problem.}
The GPU underutilization in single-GPU, data-parallel, and pipe- line-parallel training is 
caused by different reasons, and thus different optimization techniques are previously
proposed for them\,\cite{kwon2020nimble, jayarajan2019priority, pipedream19, gpipe19}.
Although the GPU underutilization in the three training methods seems to be a separate issue,
they can be formulated as a same optimization problem.
In all three training methods, we need to optimize for the GPU utilization and training throughput.
This requires carefully scheduling the operations in the training to maximize, for example,
the overlapping of the computation and communication.
Commonly for the three training methods, we formulate the problem of optimizing the execution 
of a single forward and backward propagation as following.
\begin{equation*}
\small
\begin{aligned}
 & \underset{\textstyle T: \mathbb{C} \rightarrow \mathbb{R}^{|\mathbb{C}|} }
             {\text{minimize}} \quad T(F_L) + F_L  \\
 & \text{s.t.} ~ T(\delta O_{L+1}) = 0, ~~
                  \{T(\delta W_i),\, T(\delta O_i) \} \geq T(S[\delta O_{i+1}])+S[\delta O_{i+1}], \\
 & \phantom{123} T(S[\delta O_{i}]) \geq T(\delta O_{i})+\delta O_{i}, 
                     ~~ T(S[\delta W_{i}]) \geq T(\delta W_{i})+\delta W_{i}, \\
 & \phantom{123} T(F_i) \geq T(S[\delta W_{i}])+S[\delta W_{i}], ~ T(F_{i+1}) \geq T(F_i)+F_i,
                     ~~~~\text{\small{where}}
\end{aligned}
\end{equation*}
\vspace{-3mm}
\begin{itemize}[leftmargin=0.1cm]
\item  $F_i$, $\delta O_i$, $\delta W_i$ are i'th layer's forward, output and weight gradient computation;\,they also denote their execution\,times.
\item $S[\delta O_i]$, $S[\delta W_i]$ are the synchronization of $\delta O_i$, $\delta W_i$.
These may be no-op;\,e.g. $S[\delta O_*]$ is no-op for data-parallel training.
\item $L$ is the number of layers, $\mathbb{C}$ is the set of all operations, i.e., $\{F_1, \delta W_1,$ $S[\delta W_1],  ...\}$, and $\mathbb{R}$ is the set of real numbers.
\item $T:\mathbb{C} \rightarrow \mathbb{R^{|C|}}$ is the scheduling function that determines the start time of the operations;
for example, $T(F_L)$ is the start time of the forward computation of layer $L$.
\end{itemize}
The goal is to find the function $T$ that minimizes the makespan of the executions.
Note that we start with the backpropagation and end with the next iteration's forward propagation;
the gradient computation of loss ($\delta O_{L+1}$) is scheduled at time zero and
the completion time of the last layer's forward computation ($T(F_L)+F_L$) is minimized.

This is a variant of job shop scheduling problem\,\cite{jobshop1966},
which is hard to solve accurately\,\cite{jobshopapprox1987,jobshopcomplexity1976}.
Heuristic algorithms such as list scheduling\,\cite{de1994synthesis} or HEFT\,\cite{topcuoglu2002performance} are generally used to find a good solution that satisfies the constraints.
We show in Section~\ref{sec:single-gpu} and \ref{sec:multi-gpu} that our algorithm based on
the list scheduling technique\,\cite{de1994synthesis} can find optimized kernel schedules for all three training methods (with different prioritization schemes).
Before we present the scheduling algorithms, we first describe our core scheduling technique
that allows flexible scheduling of neural network operations.

\section{Out-Of-Order BackProp}
\label{sec:ooobp}
\begin{figure}[b]
\centering
\vspace{-1mm}
\includegraphics[width=3.35in]{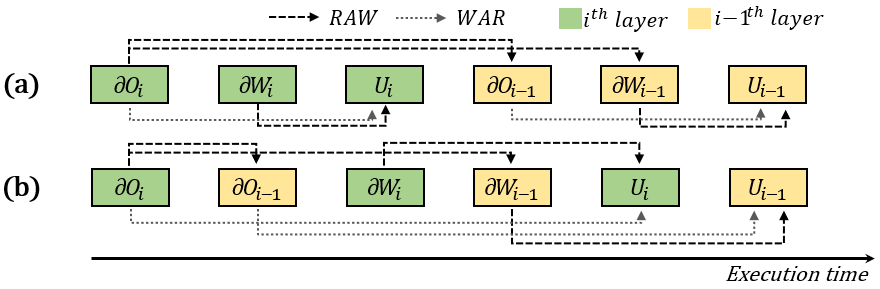}
\vspace{-2mm}
\caption{Execution timelines of backpropagation. The dependencies 
of output/weight gradient computations and weight updates\,($\delta O$,$\delta W$,\,$U$) are
shown in arrows;
(a) is conventional backpropagation and (b) is out-of-order backprop.}
\label{fig:backward_timeline}
\end{figure}

In forward propagation of DNN training, each layer computes the output 
with its input, which is the prior layer's output.
The computed outputs are stored for backpropagation.
The final output is compared with the target value to compute the loss.
In backpropagation, the computation is performed
in the backward direction to calculate the
output gradient\,($\delta O$) and weight gradient\,($\delta W$) for each layer.
The computed output gradient is used to calculate the gradients in the prior layer.
However, the weight gradient\eat{YYY sout, representing the partial derivative of the loss
with respect to the layer's weight,} is not used to compute any other gradients;
it is only used to update the layer's weight parameters.
This dependency is shown in Figure~\ref{fig:backward_timeline}\,(a).

Exploiting the dependencies of gradient computations, we propose 
{\it out-of-order backprop} (ooo backprop in short), which schedules the 
weight gradient computations out of their layout order.
In conventional backpropagation, the gradient computation and weight update
are performed in the reverse order of the layers in a network. That is,
the two gradient computations for a layer are completed before
starting the previous layer's gradient computations.
Figure~\ref{fig:backward_timeline}\,(a) shows the execution of conventional backpropagation.

Out-of-order backprop decouples the gradient computation
from the structure of a neural network, thus scheduling 
weight gradient computations and updates in a flexible manner.
As a result, more critical\,(bottleneck-causing) computations 
can be scheduled with higher priorities.
For example, when applied to pipeline-parallel training,
ooo backprop schedules all the output gradient computations of a GPU 
before its weight gradient computations so that the next GPU may promptly start
its computation.
This is shown in Figure~\ref{fig:backward_timeline}\,(b). The current GPU
computes $\delta O_i$ and $\delta O_{i-1}$, transferring the output to the next GPU
assigned with $i$$-$$2$'th and $i$$-$$3$'th layers; then it computes $\delta W_i$ and $\delta W_{i-1}$
concurrently with the next GPU computing $\delta O_{i-2}$ and $\delta O_{i-3}$\,(not shown).

When scheduling weight gradient computations, we need to 
consider both GPU utilization and memory overhead at the same time.
Because the weight gradient computation of a layer requires
the layer's input and output gradient, those values must be retained in memory until
the computation is completed.
Our scheduling algorithms in Section~\ref{sec:single-gpu} and \ref{sec:multi-gpu}
take this memory overhead into account and find efficient execution schedules
with minimal memory overhead
when applying out-of-order backprop in single- and multi-GPU training.

\section{Scheduling for Single-GPU Training}
\label{sec:single-gpu}
In single-GPU training the kernel issue/execution overhead and 
idling SMs in kernel executions cause the GPU underutilization.
This section presents our {\em multi-stream out-of-order computation} that 
applies concurrent GPU streams and ooo backprop to improve the GPU utilization
for single-GPU training.

\subsection{Multi-Stream Out-Of-Order Computation}
To mask the kernel execution overhead and GPU's idle cycles,
we propose to use two GPU streams, namely {\em main-stream} and {\em sub-stream}.
In main-stream we allocate the operations in the critical path, i.e.,
output gradient computations and forward computations of all layers; 
we set this stream's priority high in the GPU execution engine.
In sub-stream we run weight gradient computations and weight updates.
Using two GPU streams in this manner requires an additional constraint in the 
optimization problem in Section~\ref{sec:underutil_gpu}: 
$T(\delta W_i)$$\geq$$T(\delta W_j)$$+$$\delta W_j$ if $T(\delta W_i)$$\geq$$T(\delta W_j)$.

To solve this optimization problem, we use the list scheduling technique and
prioritize the co-scheduling of the kernels with highest speedup when run together.
Because the GPU execution engine dynamically determines the SM allocations for the kernels,
it is not feasible to apply fine-grained scheduling to exactly overlap 
the executions of two kernels.
\eat{(unless fine-grained synchronization is used, which is expensive)}
Hence we exercise coarse-grained control and apply region-based scheduling.
That is, we divide the forward and backward propagation into multiple regions with similar compute characteristics;
e.g. a ResNet block can be a single region (forward and backward separately) as
it consists of the same repeated convolutions.
Then for each region we co-schedule the kernels that give highest speedups.

More specifically our scheduling proceeds as following:
\begin{enumerate}[leftmargin=4mm]
\item For the possible region pairs we profile their concurrent kernel runs and record the speedups over their sequential runs.
\item We sequentially schedule the main-stream kernels.
\item For the sub-stream kernels, we compute their schedulable time intervals and regions;
then we assign the kernels to those regions as the schedulable candidates.
\item For each region, among its candidates we find the sub-stream kernel with the highest speedup for the main-stream kernels in the region.
Then we select the region-kernel pair with the highest speedup and schedule the kernel in the region.
\item We repeat step 4 until all the sub-kernels are scheduled.
\end{enumerate}

\SetKwInput{KwInput}{Input}
\SetKwInput{KwOutput}{Output}
\begin{algorithm}[t]
\small
  \caption{Multi-Region Joint Scheduling\label{algo:region-sched}}
\KwInput{$R[1..N]$: main-stream kernel schedule split into $N$ regions,
$T_{main}(R[i])$: total exec time of $R[i]$'s main-stream kernels,
$T_{sub}(k, R[i])$: exec time of the sub-stream kernel $k$ in $R[i]$.
}
\KwOutput{Sub-stream's schedule $S[1..N]$ for $N$ regions} 
 $S[i] = [ ], ~now[i] = 0 \textup{ for } i=1, ..., N$\;
 $\mathcal{U}=\{\delta W_2, ..., \delta W_L \} $; ~ $\mathcal{C}=\{R[1], ..., R[N]\}$\;
 \While {$\mathcal{U} \neq \emptyset$} {
   \ForEach {\textup{region} $R[i] \in \mathcal{C}$} {
     find $\delta W_k$ runnable at $now[i]$ w/ max speedup $p$ in $R[i]$\\
     $tmp[i] = (\delta W_k, p)$\;
   }
   $j = \arg\max_{i}\,f(i)=tmp[i][2]$\;
   $(\delta W_K, P) = tmp[j]$\;
   $S[j]\textup{.append}(\delta W_K)$;~ $\mathcal{U}{.remove}(\delta W_K)$\;
   $now[j] \leftarrow now[j]+T_{sub}(\delta W_K, R[j])$\;
   \lIf{$now[j] \geq T_{main}(R[j])$} {
     $\mathcal{C}\textup{.remove}(R[j])$
   }
 }
 \Return{S[1..N]}\;
\end{algorithm}

Note that the overhead of the profiling is minimal because it can be
performed as part of the training and also the number of regions that we use
is fairly small; in our evaluation we used eight regions for DenseNet-121.
This algorithm is a variant of list scheduling, 
which divides the timeline into multiple regions and jointly schedules for those regions altogether.
This region-based approach works well in practice because
recent neural networks often have sub-structures with similar operations such as DenseBlocks or ResNet blocks. 
Algorithm\,\ref{algo:region-sched} shows the pseudocode for step 4 and 5 above.
In the algorithm, we simulate the timeline of the regions (denoted by $now[i]$)
using the expected kernel execution time and the profiling results.
In lines 4--6 we find the sub-stream kernel that gives the highest speedup in each region; 
then we select the kernel and region with the highest speedup (lines 7--8)
and schedule the kernel in the region (line 9). We update the timeline for the scheduled region (line 10--11) and repeat the scheduling process.

With the execution schedule given by Algorithm\,\ref{algo:region-sched}, we compute its memory usage. 
If the peak memory usage is too high, we pre-schedule for $k$ regions
at the beginning of the backpropagation;
in the $k$ regions, we schedule the weight gradient computations as soon as they are ready,
and thus the peak memory usage is decreased.
Then we re-try running Algorithm\,\ref{algo:region-sched} for the remaining regions, increasing $k$ for each re-try.

\subsection{Pre-compiled Kernel Issue}

Multi-stream ooo computation reduces the idling SMs by overlapping the kernel executions. 
However, the optimization may not be effective when the kernel issue overhead is large.
If it takes too long to issue the kernels from the CPU side, executing them
on multiple GPU streams does not help to mask the idle cycles.
To mitigate the kernel issue overhead, we apply {\em pre-compiled kernel issue}.

In neural network training, the same computation graph is repeatedly executed 
and thus the same sequence of kernels are invoked over and again.
NVIDIA released CUDA Graph API\,\cite{Gray2019CUDAgraphs} that supports {\em capturing} a sequence of kernel issues and pre-compiling them into an {\em executable graph}.
We can launch the executable graph with very little overhead using CUDA Graph Launch API.
We apply this technique to reduce the kernel issue overhead so that 
our multi-stream ooo computation (with multi-region joint scheduling)
can effectively reduce the idling SMs and improve the GPU utilization.
A similar technique of collectively launching neural network kernels
has been recently used by Nimble\,\cite{kwon2020nimble};
MXNet\,\cite{chen2015mxnet} applies a similar principle but at the framework level without using CUDA Graph API.
We use the pre-compilation technique together with multi-stream ooo computation 
to maximize the GPU utilization.

\section{Scheduling for Multi-GPU Training}
\label{sec:multi-gpu}
In multi-GPU training,
the main performance overhead is  
1) parameter communication in data-parallel training and
2) pipeline stalls in pipeline-parallel training.
This section describes our scheduling algorithms for multi-GPU training.

\subsection{Data-Parallel Training}
\label{sec:data-parallel}

\begin{figure}[t]
\centering
\includegraphics[width=3.35in]{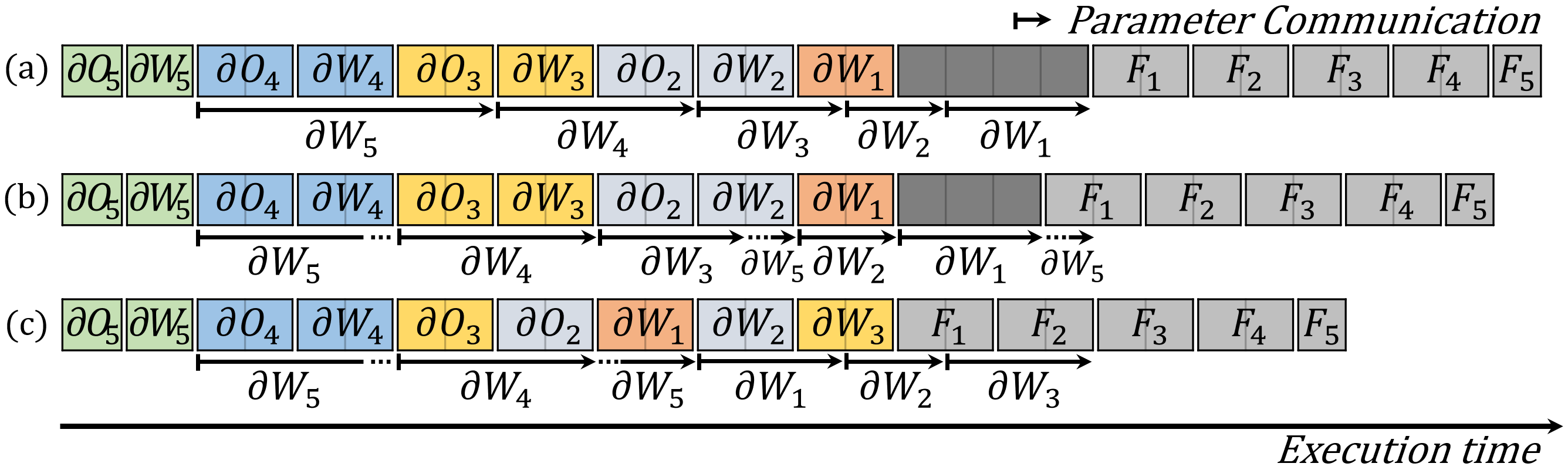}
\vspace{-8mm}
\caption{Data-parallel training of a neural network. Execution (a) uses conventional backpropagation; (b) 
prioritizes the parameter communications. Execution (c) further prioritizes the gradient computations.
 The dark gray box is GPU idle time. }
\label{fig:data-par-sched}
\vspace{-3mm}
\end{figure}

The optimization of data-parallel training is equivalent to the problem we defined in Section~\ref{sec:underutil_gpu},
if the synchronization of output gradient ($S[\delta O_*]$) is set to no-op with empty execution time.
Like the single-GPU training case, this problem is NP hard and we propose a heuristic scheduling algorithm.

In data-parallel training, the parameter synchronization overhead ($S[\delta W_*]$ in the problem definition)
is the major performance bottleneck. Figure~\ref{fig:data-par-sched} shows an example execution timeline of data-parallel training.
In conventional backpropagation (a), the parameter communication, denoted by the arrows, 
postpones the forward computations and results in GPU idle cycles (the dark gray boxes in the timeline).
If we apply the prioritized parameter communication technique that is proposed
in recent studies\,\cite{peng2019generic,jayarajan2019priority,byteps,hashemi2018tictac,li13pytorch}, 
the performance is slightly improved as shown in Figure~\ref{fig:data-par-sched}\,(b).
The communication of $\delta W_1$--$\delta W_4$ is prioritized over $\delta W_5$ (denoted by the dotted arrow),
which reduces the execution time by one unit time.
In addition to the communication prioritization, we can further improve the performance
by prioritizing the {\it computations} in the critical path.
Specifically, let us schedule the computation of $\delta W_2$ and $\delta W_3$
(in this order) after the computation of $\delta W_1$ as shown in 
Figure~\ref{fig:data-par-sched}\,(c).
Then the communication of $\delta W_1$ is masked by the computation of $\delta W_2$ and $\delta W_3$.
The training time is reduced by three unit times, improving the performance by 16\% compared to conventional backpropagation in (a) and by 12\% compared to the prioritized communication in (b).
Note that we can obtain this optimal schedule by reversing the computation of $\delta W_1$ -- $\delta W_3$;
for this example, we can obtain another optimal schedule if we reverse
the computation of $\delta W_1$ -- $\delta W_4$.

To find the optimal execution schedules such as Figure~\ref{fig:data-par-sched}\,(c),
we need to design a list scheduling algorithm
and prioritize the computations in the critical path.
However, since parameter synchronization is the dominant factor of the performance,
we achieve (mostly) the same effect by simply advancing the gradient computations
upon which the critical synchronizations depend.
Thus we design a heuristic algorithm, namely reverse first-$k$ scheduling,
that reversely orders the weight gradient computations of first $k$ layers.
Because those $k$ layers are computed earlier in the forward propagation than the other layers,
the synchronization of their weight gradients are the critical operations;
hence prioritizing their gradient computations shortens the critical path and minimizes
the total execution time.
For example, the executions in Figure~\ref{fig:data-par-sched}\,(c) is equivalent to the result of 
applying reverse first-$k$ scheduling with $k$=3.
To find the optimal $k$, we profile the executions of the schedules
with multiple $k$ values.

\SetKw{KwDownto}{down\,to}
\SetKw{Break}{break}

\let\oldnl\nl
\newcommand{\nonl}{\renewcommand{\nl}{\let\nl\oldnl}}
\begin{algorithm}[tb]
\small
\caption{Reverse First-{\it K} Scheduling\label{algo:rev-k}}
\KwInput{$k$: number of layers to reschedule, $MXM$: max memory.}
\KwOutput{Optimized schedule $D$} 
$max_k = \arg\max_{j}f(j)=M_{fwd} - \sum_{i=j+1}^{L} M(\delta O_i)+ \sum_{i=1}^{j}M(\delta W_i)$\\
\nonl  \phantom{01234567890123456789012} \text{\small{s.t.}} {\small $f(j) < MXM$}\;
$k \leftarrow \min(max_{k}, k)$\;
\For {$i\leftarrow L$ \KwDownto $1$} {
  \lIf{$i>k$} {$D.append(\delta W_i)$}
  $D.append(\delta O_i)$\;
}
\lFor {$i\leftarrow 1$ \KwTo $k$} {
  $D.append(\delta W_i)$ }
\Return{D}\;
\end{algorithm}

Algorithm~\ref{algo:rev-k} describes reverse first-$k$ scheduling.
It reorders the backpropagation portion of the training.
In lines 1--2, it adjusts the value of $k$ to satisfy the given memory constraint;
function $M()$ returns the amount of temporary memory that is used by the computation.
In lines 3--5, it schedules the gradient computations from layer $L$ down to layer $1$
in the same way as conventional backpropagation except for the
weight gradient computations of layer 1 to layer $k$\,(line 4).
Lastly it schedules the weight gradient computations of layer $1$ to layer $k$ in the reverse order\,(line 6).

If the optimal value of $k$ is given, Algorithm~\ref{algo:rev-k} effectively prioritizes the
critical synchronizations. If $k$ is too small, the synchronizations are not fully masked; if it is too large, the network bandwidth may be underutilized.
With profiling the executions, we can find the optimal $k$ that gives the fastest execution.
List scheduling, on the other hand, does not need to find such optimal values 
but it requires the execution times of the parameter synchronizations.
Because it may not be easy to estimate the synchronization time,
reverse first-$k$ scheduling is more effective and suitable in practice.

\subsection{Pipeline-Parallel Training}
\label{sec:pipeline-par}
In pipeline-parallel training, each GPU is assigned with
a subset of layers and executes the computations for the assigned layers.
The optimization of pipeline-parallel training is equivalent to 
the optimization problem in Section~\ref{sec:underutil_gpu},
if we set $S[\delta W_*]$ to be no-op.

\begin{figure}[b]
\centering
\includegraphics[width=3.35in]{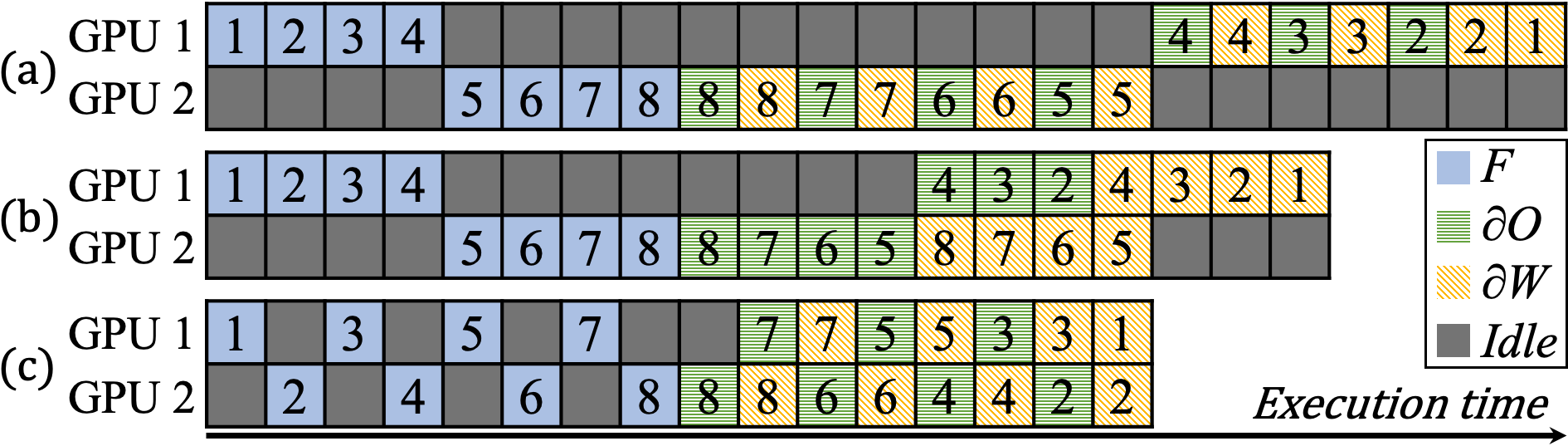}
\vspace{-5mm}
\caption{Pipeline-parallel training of an 8-layer neural network; (a) is conventional execution, (b) applies gradient fast-forwarding, and (c) further applies modulo allocation.}
\vspace{-2mm}
\label{fig:pipe_compare}
\end{figure}

As in the other training methods, we can improve the performance of pipeline-parallel training
by prioritizing the critical computations.
As such, we propose {\em gradient fast-forwarding}, which prioritizes the executions of
the output gradient computations over those of the weight gradient computations in the GPU's allocated layers.
This way, the weight gradient computations in one GPU
may overlap with the output gradient computations in the next GPU. 
Figure~\ref{fig:pipe_compare}\,(a) and (b) compare the execution timeline of training with two GPUs without and with the fast-forwarding.
In (a), cross-layer model parallelism is applied and only a single GPU is used at a time;
in (b) with gradient fast-forwarding, the computations of GPU\textsubscript{1} run concurrently with the weight gradient computations of GPU\textsubscript{2}. 
The total execution time is reduced from 23 to 19 unit times
resulting in 21\% speedup.
For the simplicity of the presentation, we describe our optimizations
without the micro-batch technique\,\cite{gpipe19,pipedream19}, but our optimizations work well with 
micro-batches as we discuss at the end of this section. 
In Section~\ref{sec:eval-pipepar}, we show the effect of our optimizations over micro-batches
in the experimental results and also in the timeline analysis (Figure\,\ref{fig:pipeline_timeline}).

\noindent
{\bf Modulo allocation.}
We further speed up pipeline-parallel training 
by efficiently partitioning a neural network and allocating the partitions to GPUs.
In conventional pipeline-parallel training, consecutive layers of a neural network
are assigned to a same GPU to minimize inter-GPU communication overhead\,\cite{gpipe19, pipedream19}.
Contrary to this conventional practice, we propose {\em modulo layer allocation} 
that may increase the inter-GPU communication.
That is, when training with $n$ number of GPUs, we allocate $l$'th layer of the neural network to GPU\textsubscript{$l$ mod $n$};
GPU\textsubscript{i} computes for layer $l_i$, $l_{i+n}$, $l_{i+2n}$, etc.
While it may increase the communication overhead, this modulo allocation gives much higher GPU utilization than the conventional allocation scheme, as we elaborate below.

Consider the training of the eight-layer neural network in Figure~\ref{fig:pipe_compare}\,(b).
With gradient fast-forwarding, half of the backpropagation runs concurrently,
but the two GPUs are idle for the other half of the backpropagation.
If modulo allocation is further applied, 
both GPU\textsubscript{1} and GPU\textsubscript{2} are utilized for more than
90\% of the backpropagation as shown in the timeline in Figure~\ref{fig:pipe_compare}\,(c).
Compared to the conventional execution in (a), the execution in (c)
takes only 16 unit times, which is 1.44$\times$ speedup.

Moreover, this technique reduces the (additional) memory overhead.
Gradient fast-forwarding, if used with the conventional layer allocation scheme, 
requires storing output gradients in memory for the delayed weight gradient computations.
Modulo layer allocation, however, hands over each layer's output gradient 
to the next GPU and immediately computes its weight gradient.
Because the output gradient is discarded right after this computation, it is 
unnecessary to store more than one output gradient in each GPU.

One drawback of modulo allocation is the increased inter-GPU communication.
In our evaluation, we measured this overhead in low and high-bandwidth network.
The details are discussed in Section~\ref{sec:eval-pipepar} 
but the short summary is that even with the communication overhead,
our optimizations achieve a large performance gain over the leading edge pipeline-parallel systems (GPipe and PipeDream).

\noindent
{\bf Micro-batches.}
The technique of splitting mini-batches into micro-batches
is generally used in pipeline-parallel training\,\cite{gpipe19,pipedream19}.
With this technique, GPUs can concurrently compute with
different micro-batches for their allocated layers.
For example, GPU\textsubscript{1} computes with 
batch B\textsubscript{1} and hands over the output to GPU\textsubscript{2};
then GPU\textsubscript{1} computes with the next micro-batch B\textsubscript{2},
concurrently with GPU\textsubscript{2} working on B\textsubscript{1}.
This way, the overall GPU utilization is improved.
Our techniques work well with micro-batches to further 
improve the training throughput as shown in  
our evaluation (Section~\ref{sec:eval-pipepar}).

\section{Implementation in TensorFlow}
\label{sec:impl}

We implemented our optimizations in TensorFlow\,(v2.4) and its optimizing compiler XLA.
To implement out-of-order backprop, we eliminated the use of \texttt{tf.group}
that puts the weight and output gradient computations
of a layer into a single node in the computation graph. 
By putting them in separate nodes, we remove the unnecessary dependencies 
for those gradient computations.

\begin{figure}[t]
\centering
\includegraphics[width=3.3in]{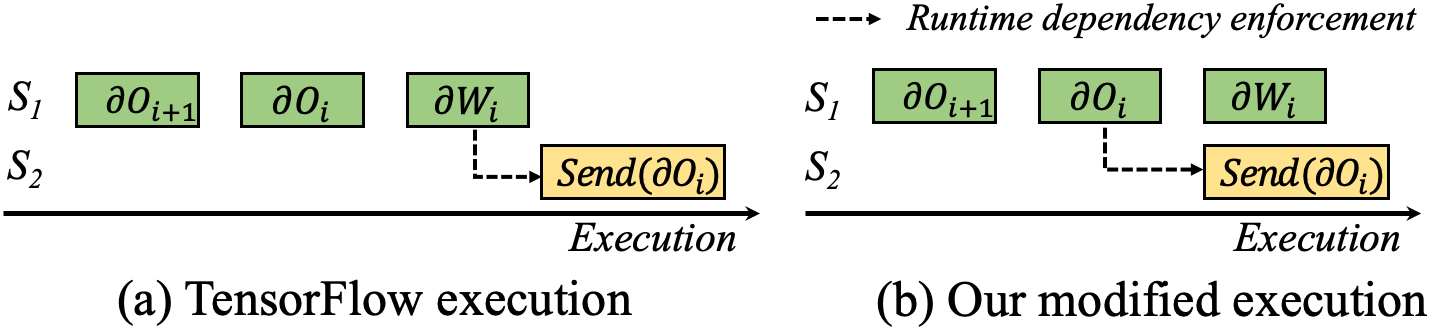}
\vspace{-1mm}
\caption{Scheduling I/O operations in TensorFlow.
\label{fig:send_order}}
\vspace{-1mm}
\end{figure}

To maximize the overlapping of computation and communication, 
we fixed the runtime enforcement of operation dependencies in TensorFlow.
Figure~\ref{fig:send_order} shows a part of an execution timeline with 
computations in GPU stream $S_1$ and synchronization in stream $S_2$.
When issuing {\it Send}$(\delta O_i$) in stream $S_2$,
TensorFlow enforces the dependency using CUDA event/stream APIs
and simply makes it to be executed after {\it any} operation just issued in $S_1$.
Because {\it Send}$(\delta O_i$) is issued after 
$\delta O_{i+1}$, $\delta O_i$, and $\delta W_i$ are issued (in this order),
{\it Send} is forced to execute after the computation of $\delta W_i$.
We fix this and make {\it Send} to execute after $\delta O_i$'s computation,
which makes {\it Send} to overlap with $\delta W_i$'s computation.

Moreover, we implemented an efficient support for an auxiliary GPU stream in TensorFlow.
Although TensorFlow has preliminary implementation for multiple GPU streams (which is disabled by default\,\cite{disable_multi_stream}),
it uses much more memory compared to the single-stream executions.
That is, in single-stream executions, temporary memory that is solely accessed by a kernel 
may be immediately reclaimed and used by other kernels as soon as the kernel is {\it issued} 
(before its execution finishes).
However, with the (generic) multi-stream support in TensorFlow, 
reclaiming temporary memory in this manner is not feasible
because of the complex dependencies between kernels in different streams.
Rather than using the generic but expensive multi-stream support in TensorFlow, 
we implemented a light-weight version that supports only one additional stream (i.e., sub-stream),
for running the weight gradient computations. 
We also assigned a separate memory allocator for the temporary memory used by the sub-stream kernels.
We used NVIDIA's event/stream APIs to enforce the dependency between 
the main-stream and sub-stream kernels.

Our prototype is based on TensorFlow, but all our techniques can be implemented in 
MXNet or PyTorch, by modifying, e.g., PyTorch's autograd engine
that dynamically constructs the backward graphs.

\begin{table}[tbh]
\caption{Models, datasets, and evaluation setup. 
}
\small
\centering
\setlength{\tabcolsep}{2pt} 
\renewcommand{\arraystretch}{1} 
\begin{tabular}{c|c:c:c}
\specialrule{.1em}{.05em}{.05em}
  \textbf{Training Method} &
  \textbf{Model} &
  \textbf{Dataset} &
  \textbf{GPU} \\
\specialrule{.1em}{.05em}{.05em}
\hline
 Single GPU       & DenseNet-\{121,169\} &  CIFAR100  &  Titan\,XP  \\
 Training         & MobileNet V3 Large &  ImageNet  &  V100              \\ 
                  & ResNet-\{50,101\} &   &                    \\ \hline
 Data-Parallel    & DenseNet\{121,169\} &     & Titan\,XP $\times$ 8 \\
 Training         & MobileNet V3 Large &  ImageNet  & P100 $\times$ 20    \\
                  & ResNet-\{50,101,152\} &   & V100 $\times$ 48    \\ \hline 
Pipeline-Parallel & RNN\,(16 Cell), FFNN  &   IWSLT    &                   \\
 Training         & BERT-\{12,24,48\}& MNLI & V100 $\times$ 36  \\
                  & GPT-3 (Medium) &  OpenWebText  &                   \\  \hline
\end{tabular}
\label{tab:models}
\end{table}

\section{Evaluation}
\label{sec:eval}

In this section, we present a comprehensive evaluation of
our scheduling algorithms. 
We evaluate our optimizations with twelve neural networks
and five public datasets on a single and multiple GPUs.
Because our optimizations do not change the semantics of neural network training,
we only evaluate the training throughput and the memory overhead.
The details of the evaluation are described later,
but the short summary is that out algorithms with 
out-of-order backprop effectively improves
the performance of single-GPU training as well as
data- and pipeline-parallel training.
Compared to the respective state of the art systems,
we speed up the throughput by 1.03--1.58$\times$ for single-GPU training,
by 1.10--1.27$\times$ for data-parallel training, and by
1.41--1.99$\times$ for pipeline-parallel training.

\subsection{Experimental Setup}

\noindent
{\bf Models and datasets.}
Table~\ref{tab:models} describes the twelve neural networks and five datasets
that are used for the evaluation. 
These models are state of the art neural network models that are commonly 
used in computer vision and natural language processing (NLP).
DenseNet-\{121, 169\}, MobileNet, and ResNet-\{50,101,152\} are established 
CNN models in computer vision.
DenseNet has {\it growth rate} as its hyper parameter (denoted by $k$);
we set $k$=12, 24, and 32, which are the same as those used by the authors\,\cite{huang2017densely}.
MobileNet also has a hyper parameter, namely {\it multiplier} (denoted by $\alpha$);
we use $\alpha$=0.25, 0.5, 0.75, 1, which are also the same as those used by its authors\,\cite{howard2017mobilenets}.
For the language processing, we used Recurrent Neural Networks (RNNs), BERT-\{12,24,48\}, and GPT-3 (Medium) as they are the representative NLP models.
We also experimented with a simple feed forward neural network (FFNN).
We used public datasets that are widely used to evaluate CNNs (CIFAR100\,\cite{krizhevsky2009learning} and ImageNet\,\cite{russakovsky2015imagenet})
and language models (IWSLT\,\cite{Cettolo2015TheI2}, MNLI\,\cite{williams2018broad}, and OpenWebText\,\cite{Gokaslan2019OpenWeb}). 

\noindent
{\bf GPUs and interconnects.}
We used NVIDIA Titan XP, P100, and V100 GPUs for the training as shown in Table~\ref{tab:models}.
Titan XP GPUs are installed on a small cluster of eight machines.
The machines have Intel Xeon E5-2620 v4 running at 2.1GHz and 64GB of DRAM.
Each machine has a single Titan XP GPU connected via PCIe 3.0 x16;
the machines are connected via 10Gb Ethernet.
P100 GPUs are deployed on a cluster of twenty machines,
each containing one P100 GPU connected via PCIe 3.0 x16;
the machines have Intel Xeon E5-2640 v3 running at 2.6GHz and 32GB of DRAM.
The machines are connected via 20Gb Ethernet.
V100 GPUs are installed on a public cloud (Amazon AWS). The instance types for the 
evaluation are shown in Table~\ref{tab:clusters}, which summarizes the cluster settings.
We used the two private clusters described earlier and two public clusters on AWS.
The AWS instances have Intel Xeon E5-2686 v4 running at 2.3GHz.
The instances in Pub-A cluster have 244GB of DRAM and four V100 GPUs and
the instances in Pub-B cluster have 488GB DRAM and eight V100 GPUs.
We used up to forty eight V100 GPUs in Pub-A cluster and up to thirty six V100 GPUs in Pub-B cluster.

\begin{table}[t]
\caption{GPU cluster settings.}
\small
\centering
\begin{tabular}{cccc}
\hline
\textbf{Cluster} & \textbf{Instance} & \textbf{GPUs} &  \textbf{Interconn.} \\
\textbf{Name}    & \textbf{Type}   & \scriptsize{(\# per node$\times$node \#)} 
                                   & \scriptsize{inter-GPU, inter-node} \\
\hline
Priv-A & Private Cluster & Titan\,XP (1$\times$8) & PCIe, 10Gb \\
Priv-B & Private Cluster & P100 (1$\times$20) & PCIe, 20Gb \\
Pub-A & AWS p3.8xlarge & V100 (4$\times$12) & NVLink, 10Gb \\
Pub-B & AWS p3.16xlarge & V100 (8$\times$5) & NVLink, 25Gb \\
\hline
\end{tabular}
\label{tab:clusters}
\vspace{-5mm}
\end{table}

\noindent
{\bf Other training settings.}
We train with the batch sizes that are used by the authors of the models;
we also tested with the maximum batch sizes on the GPUs.
For DenseNet, MobileNet, and ResNet models we set the batch size per GPU
to be 32, 64, 96, and 128 for CIFAR100 and ImageNet\,\cite{resnet2016, huang2017densely, howard2017mobilenets}.
The maximum global batch size that we used is 6,144 for ResNet-50 with 48$\times$V100 GPUs.
To evaluate BERT and GPT-3, we set the batch size to be 96 for the fine-tuning\,\cite{shoeybi2019megatron,lan2019albert,yang2019xlnet,dai2020funnel};
for the pre-training, we set the batch size to be 512--1872 for BERT and 96--216 for GPT-3,
which is similar to commonly used batch size for pre-training the models\,\cite{shoeybi2019megatron,he2020deberta,liu2019roberta,sun2020mobilebert,song2020lightpaff,gehman2020realtoxicityprompts}.

We trained the models with multiple optimizers (SGD, momentum, RMSProp, and Adam optimizers)
and report the throughput with momentum optimizer as training with other optimizers show similar trend.
For BERT and GPT, we use Adam optimizer which is used by its authors\,\cite{devlin2018bert, brown2020language}.
To measure the training throughput, we start the training and wait 
a few epochs for the training to warm up.
Then we measure the throughput by taking the average over ten iterations; we repeat this ten times and report the average and the standard error of the ten runs.

For the memory evaluation, we set TensorFlow's \texttt{allow\_growth} flag as \texttt{true}
to compactly allocate the required memory. 
We used \texttt{nvidia-smi} to measure the memory usage; we also investigate 
and report the memory allocation of TensorFlow's \texttt{bfc\_allocator}.

\subsection{Evaluation of Single-GPU Training}
\label{sec:eval-single}

We first measured the speedup in single-GPU training brought by
multi-stream out-of-order computation and pre-compiled kernel issue.
As the optimizations may incur memory overhead, we also 
report the additional memory usage.

\vspace{1mm}
\noindent
{\bf Training throughput.} We measure the throughput of training DenseNet, MobileNet, and ResNet in Table~\ref{tab:models} with CIFAR100 and ImageNet. 
We evaluate TensorFlow 2.4 XLA (the baseline) and XLA with our two optimizations.
For comparison we also evaluate Nimble\,\cite{kwon2020nimble}, a state of the art deep learning execution engine based on PyTorch's JIT compiler;
Nimble is reported to largely outperform PyTorch\,\cite{paszke2019pytorch}, TorchScript\,\cite{torchscript}, TVM\,\cite{chen2018tvm}, and TensorRT\,\cite{tensorrt}.

\begin{figure}[b]
\centering
\includegraphics[width=3.35in]{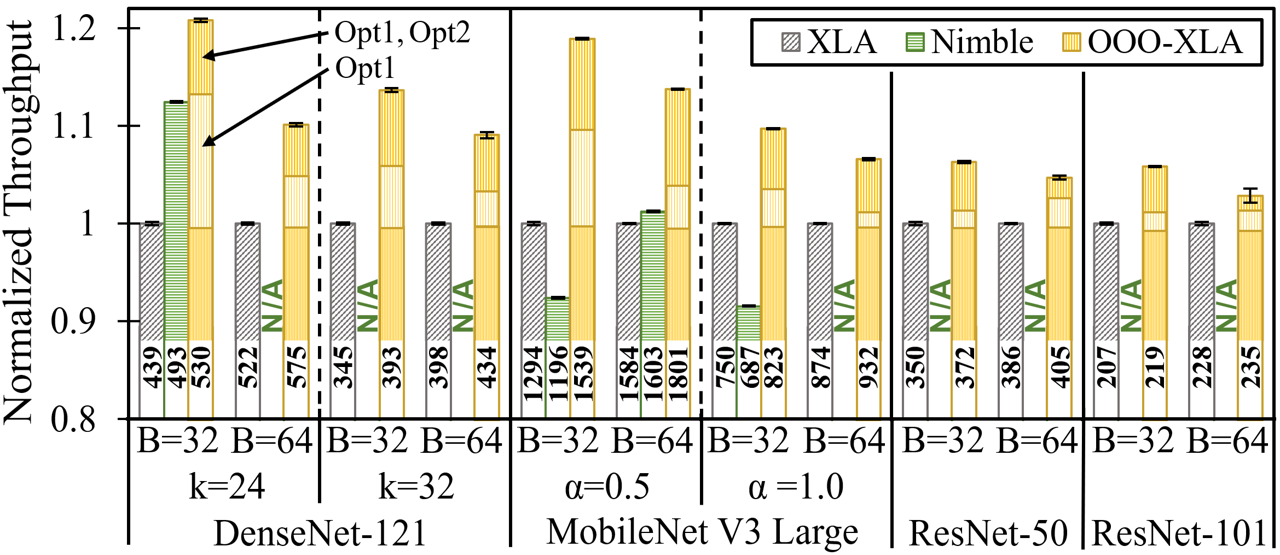}
\vspace{-2mm}
\caption{Training throughput normalized by that of XLA for batch size 32, 64.
Opt1 is pre-compiled\,kernel\,issue; Opt2 is multi-stream ooo computation.
The numbers above x-axis are the actual throughput. N/A means out of memory error.
\eat{Throughput normalized by XLA. 
B:batch size, Opt1:pre-compiled\,kernel\,issue,\,Opt2:multi-stream computation. 
The numbers above x-axis are the actual throughput.}
\label{fig:single-gpu-throughput}}
\end{figure}

Figure~\ref{fig:single-gpu-throughput} shows the training throughput of the models 
on NVIDIA V100 normalized by those of the baseline (XLA).
The numbers above the x-axis are the actual throughput (images per seconds).
For the models and batch sizes in the figure,
our optimized training (denoted by OOO-XLA) improves the throughput of XLA
by 1.09--1.21$\times$ for DenseNet-121,
by 1.07--1.19$\times$ for MobileNet, and
by 1.03--1.06$\times$ for ResNet. 
The maximum performance gain by OOO-XLA over XLA is (not shown in the figure)
1.54$\times$ for DenseNet-121 ($k$=12 and batch=32) and
1.58$\times$ for MobileNet ($\alpha$=0.25 and batch=32).
Compared to Nimble, our optimized training  runs faster
by maximum 1.35$\times$ (DenseNet:$k$=12 and batch=32, not shown) and minimum 1.07$\times$ (DenseNet-121: $k$=24 and batch=32); 
Nimble ran out of memory for most of the models with 64 batches (denoted by N/A).
When we examined the speedup by multi-stream ooo computation separately, 
it gives minimum 1--2\% (ResNet-\{50,101\}, batch=64) and maximum 15\% (MobileNet: $\alpha$=0.25 and batch=32, not shown) speedup over XLA with our pre-compiled kernel issue.

With 128 batch size, Nimble ran out of memory for all the tested models;
XLA and OOO-XLA ran out of memory for most of the DenseNet and ResNet models.
For MobileNet, OOO-XLA runs 1.04--1.09$\times$ faster than XLA with 128 batches.
In Titan XP, the three systems (XLA, Nimble, and OOO-XLA) ran out of memory for all the models with 128 batch sizes;
with 32 and 64 batch sizes, the performance gain of OOO-XLA is similar to that of V100. 

In summary, OOO-XLA runs 1.03--1.58$\times$ faster than XLA 
over all evaluated GPUs and models with the average speedup of 1.18$\times$.
Compared to Nimble, OOO-XLA runs 1.0--1.55$\times$ faster
in all experiments with the average speedup of 1.28$\times$.

\begin{figure}[tb]
\centering
\includegraphics[width=3.4in]{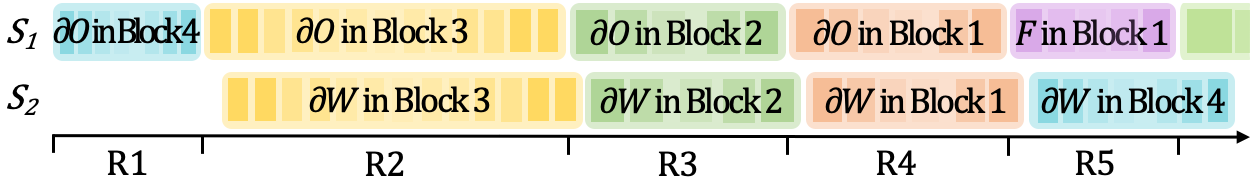}
\vspace{-3mm}
\caption{Execution timeline for DenseNet-121 with main-stream\,(S\textsubscript{1}) and sub-stream\,(S\textsubscript{2}). The kernels of the same DenseBlock are encoded with the same color.
R1--R5 are the scheduling regions, which are mapped to the DenseBlocks.\label{fig:multi-stream-ooo}}
\vspace{-1mm}
\end{figure}

\noindent
{\bf Memory overhead.} 
In all the experiments for single-GPU training, we set the memory constraint to be 1.1$\times$ of the conventional execution; i.e., OOO-XLA may use 10\% more memory 
for the reordered computations.
For all the models, the peak memory usage is only increased by maximum 0.1\%,
which is the case for DenseNet-121.
For this model, the weight gradient computations in DenseBlock-4 are delayed to
run with the forward computations in DenseBlock-1 as shown in Figure~\ref{fig:multi-stream-ooo}.
Because of the delayed weight gradient computations, the intermediate tensors need
to be stored in memory.

\begin{figure}[t]
\centering
\includegraphics[width=3.35in]{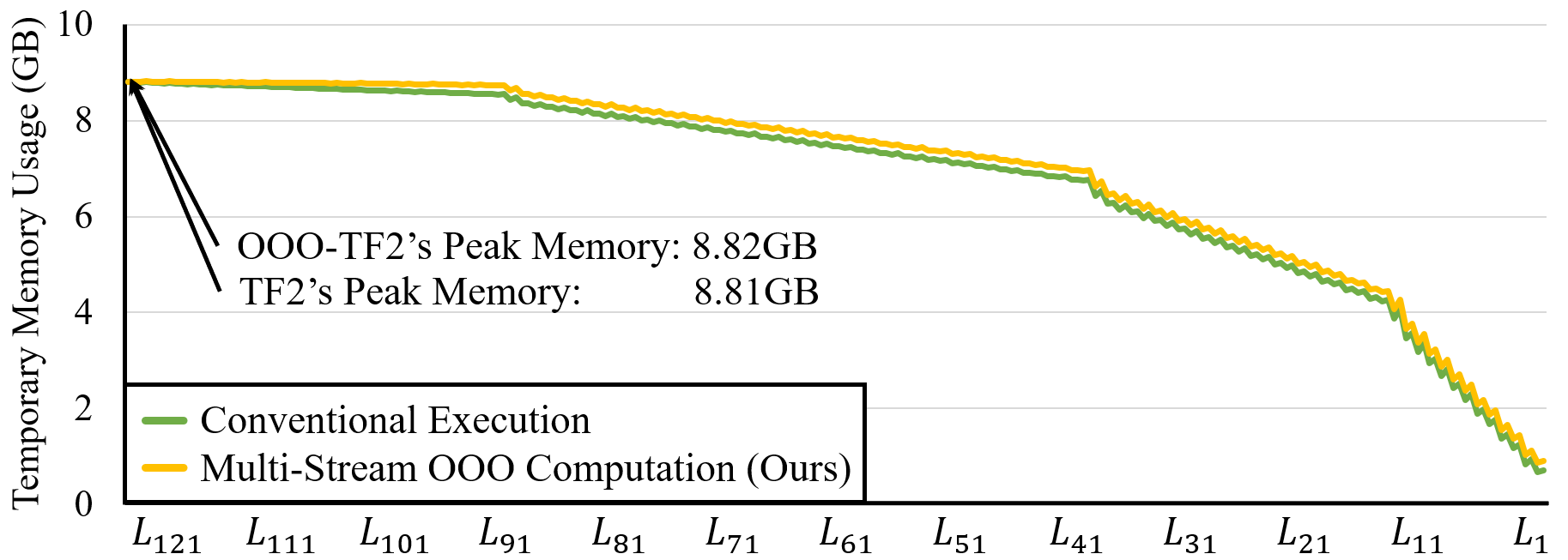}
\vspace{-2mm}
\caption{Memory usage for training DenseNet-121.\eat{, growth k=32, batch=64}}
\label{fig:single-gpu-memory-timeline}
\vspace{-3mm}
\end{figure}

This memory overhead is illustrated in Figure~\ref{fig:single-gpu-memory-timeline} 
which compares the memory usage of conventional backpropagation (green line)
and our reordered execution (yellow line).
The figure aligns the memory usage by the output gradient computations; 
that is, the memory usage of $m$ at layer $L_i$ means
that the execution uses temporary memory of $m$ when 
computing the output gradient of $L_i$.
Due to the delayed computations, our execution uses maximum 200MB more memory
than the conventional execution (in $L_{91}$--$L_{1}$ of the timeline). However, 
the peak memory is only increased by 10MB (or 0.1\%) in the beginning of the backpropagation as shown in the figure.
For the other models, the weight gradient computations of the last layers 
run concurrently with the corresponding output gradient computations, 
hence no additional memory is used.

\noindent
{\bf Discussion.} 
The kernel issue overhead is eliminated by pre-compiled kernel issue;
then multi-stream ooo computation further reduces the kernel execution overhead and the idling SMs during kernel executions. 
We examined the effect of the latter in more detail.
Specifically we look into the training of DenseNet-121 and MobileNet
where the speedup by multi-stream computation is the highest.
As shown in Figure~\ref{fig:multi-stream-ooo}, 
the weight gradient computations in DenseBlock-1--4 are reordered and executed in sub-stream.
We examine the execution of the region R2 and R5 in the figure 
because multi-stream computation gives the minimum (6\%) and maximum (10\%) speedup for the two regions respectively.

For R2, the output gradient ($\delta O$) kernels in DenseBlock-3 are running with 
the weight gradient ($\delta W$) kernels in DenseBlock-3.
More than thirty percent of the $\delta O$ kernels in R2 have the same number of thread blocks
as the SM capacity, hence the SMs are running at their maximum thread block capacity.
Hence, the performance gain by having the sub-stream kernels
is limited to reducing the kernel execution overhead.
When we sum up this overhead in R2, it totaled to about 5\% of 
R2's execution time; this is similar to the 6\% speedup achieved by multi-stream computation.

In contrast, the main-stream kernels in R5 have much larger number of
thread blocks than the SM's capacity; for the sub-stream kernels ($\delta W$ in DenseBlock-4),
half the kernels run with the same 448 thread blocks even though the SMs are 
capable of running 1,520 of them.
By running those $\delta O$ and $\delta W$ kernels concurrently, 
we provide the opportunity to make most of the SM resources and achieve 10\% speedup.

\vspace{-2mm}
\subsection{Evaluation of Data-Parallel Training}
\label{sec:eval-datapar}

\begin{figure}[tbh]
\centering
\includegraphics[width=3.35in]{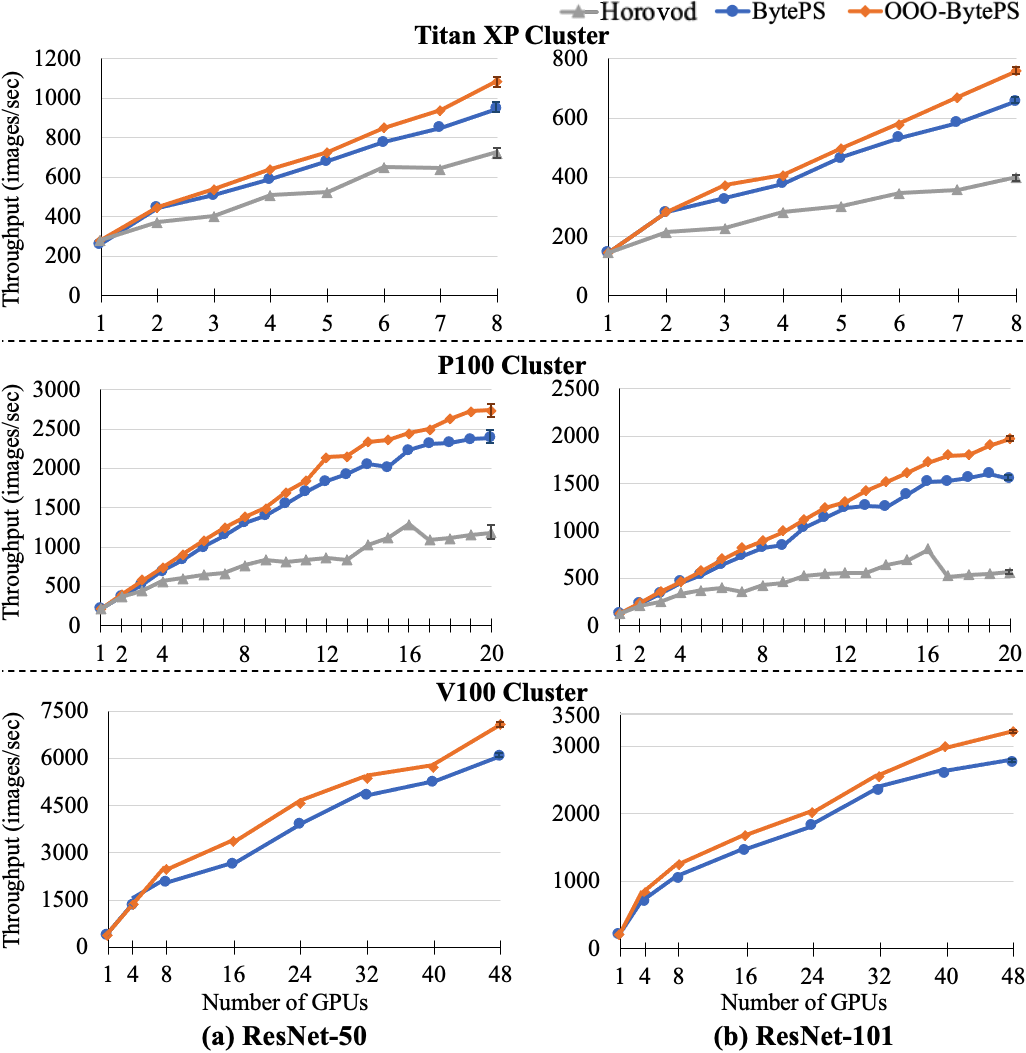}
\vspace{-5mm}
\caption{Throughput for the data-parallel training.}
\label{fig:data-par}
\vspace{-3mm}
\end{figure}

Now we evaluate our {\revsched} for data-parallel training.
We use three GPU clusters for the evaluation -- a small cluster of 8$\times$Titan XP,
a larger cluster of 20$\times$P100, and a public cloud of 48$\times$V100 (Pub-A of Table~\ref{tab:clusters}).
The settings of the GPU clusters are described in Table~\ref{tab:clusters}.
For the evaluation of data-parallel training, we trained the
DenseNet, MobileNet, and ResNet models with ImageNet.
We measured the performance improvement and memory overhead
of {\revsched}. 
The baseline for this evaluation is BytePS, the state of the art parameter-communication system for distributed training\,\cite{byteps};
BytePS is reported to be faster than other communication prioritization
systems\,\cite{jayarajan2019priority, hashemi2018tictac,byteps,peng2019generic,li13pytorch}.
We implemented our scheduling optimizations on BytePS 
and measured its training throughput with and without our {\revsched}.
For a subset of the experiments, we also evaluate Horovod,
an efficient framework for decentralized distributed training.
We do not evaluate the performance of other asynchronous training algorithms\,\cite{adpsgd2018, ssp2013, easgd2015}
as they change the training semantics.
Note that out-of-order backprop can be used with those asynchronous training algorithms to further improve their performance.

\noindent
{\bf Training throughput.} 
The results of the evaluation for the ResNet models are shown in Figure~\ref{fig:data-par};
(a) is the results for Titan XP cluster, (b) is for P100 cluster, and (c) is for V100 cluster on AWS.
The figure shows the training throughput with 1 to 8, 20, or 48 GPUs for each cluster.
For Titan XP and P100 cluster, we used the maximum 64 per-GPU batch size commonly for the two models; 
the maximum global batch size is 512 for Titan XP cluster and 1280 for P100 cluster.
For V100 cluster, we used 128 per-GPU batch for ResNet-50 and 96 per-GPU batch for ResNet-101;
the maximum global batch size is 6144 for ResNet-50 and 4608 for ResNet-101.
In Titan XP cluster, our {\revsched} (denoted by OOO-BytePS) is up to 15.3\% faster than BytePS and up to 89\% faster than Horovod (maximum speedup for ResNet-101 on 8 GPUs).
In P100 cluster, OOO-BytePS is up to 27.1\% faster than BytePS and 3.5$\times$ faster than Horovod both for ResNet-101 on 20 GPUs.
In V100 cluster, we achieve up to 26.5\% speedup for ResNet-50 on 16 GPUs and
up to 19.8\% speedup for ResNet-101 on 8 GPUs. 
In summary, OOO-BytePS achieves 1.1--1.27$\times$ speedup over BytePS
with 16--48 GPUs on P100 and V100 clusters.
For smaller models (DenseNet and MobileNet), we achieved up to 10\% and 5.3\% speedup respectively.

\begin{figure}[t]
\centering
\includegraphics[width=3.35in]{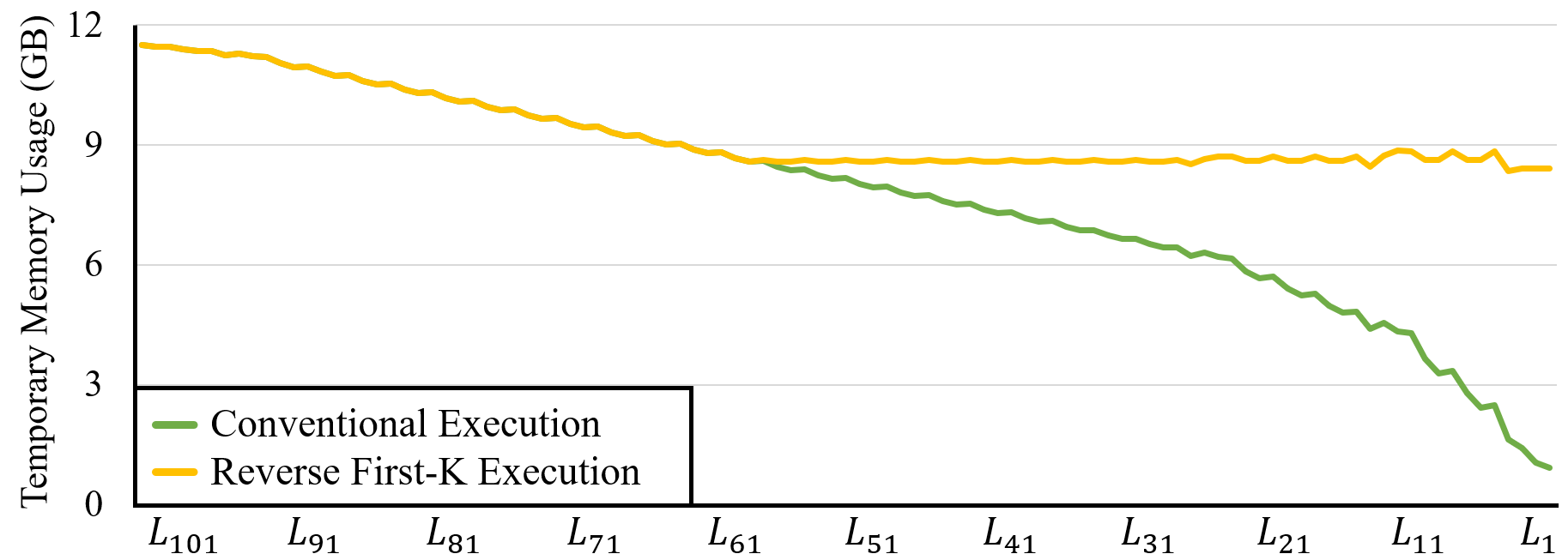}
\vspace{-2mm}
\caption{Memory usage for training ResNet-101.\eat{, batch=64}}
\label{fig:multi-gpu-memory-timeline}
\vspace{-3mm}
\end{figure}

\noindent
{\bf Memory overhead.} 
Our {\revsched} may increase the memory usage for delaying the $k$-1 weight gradient computations.
However, because the memory pressure is already reduced by running the gradient computations of
last $L$-$k$ layers (where $L$ is the total number of layers), reordering the 
computations of the first $k$ layers does not usually increase the peak memory usage.
For example, Figure~\ref{fig:multi-gpu-memory-timeline} compares the memory overhead of the conventional execution (green line) and our optimized execution (yellow line)
for ResNet-101. For our execution, first fifty five gradient computations are reversed ($k$=55) and thus  their input tensors are retained in memory increasing the memory pressure.
The memory overhead, however, does not increase but remains similar
because of the batch normalization operation, for which
the two gradient computations are performed by the same cuDNN API function 
and thus their weight gradient computations are executed in-order. 
In all our experiments with the CNN models, the additional memory overhead by {\revsched} is less than
1\% of the total memory of the conventional backpropagation execution.

\noindent
{\bf Discussion.}
To understand the performance gain of {\revsched}
we examine the training of ResNet-50 with 16$\times$V100 GPUs in Pub-A cluster.
The performance bottleneck is the first layer's weight synchronization,
which is required at the beginning of the forward computation.
With sixteen GPUs, BytePS takes 350$ms$ for this synchronization.
Thanks to the communication prioritization, the latency is much shorter than 
those reported by others using similar network settings\,\cite{Ko2021An},
but it is still substantial compared to the computation time of 380$ms$.
Our scheduling algorithm reverses first 45 layers' weight gradient computations
to overlap $\delta W_1$'s synchronization with the $\delta W_2$--$\delta W_{45}$'s computations.
The execution time of $\delta W_2$--$\delta W_{45}$ is 85$ms$, which reduces the 350$ms$ communication time 
to 265$ms$.
In addition, scheduling $\delta W_2$--$\delta W_{20}$ early in the timeline makes it possible for their
synchronization to overlap with part of the backward and forward computation, the effect of which is 65$ms$ of more overlapping.
Thus the communication overhead is reduced to 200$ms$, achieving total 27\% speedup.

\vspace{-2mm}
\subsection{Evaluation of Pipeline-Parallel Training}
\label{sec:eval-pipepar}

We evaluated pipeline-parallel training with 
NLP models -- RNN, BERT-\{12,24,48\}, and GPT-3.
We evaluated their fine-tuning and pre-training.
Fine-tuning is evaluated on four V100 GPUs;
pre-training is evaluated on up to thirty six V100 GPUs (Pub-B in Table~\ref{tab:clusters}).

We measured the speedup of our two optimizations, i.e., gradient fast-forwarding and modulo allocation.
The baseline of the experiments is GPipe, which is a state of the art pipeline-parallel training system.
We also evaluate PipeDream\,\cite{pipedream19}, another leading edge pipeline-parallel training system.
PipeDream, however, applies {\it weight stashing} that brings about parameter staleness and thus changes the semantics of the training.
Hence we report its performance as reference points only.
For a subset of the experiments, we evaluated Dapple\,\cite{fan2021dapple}, a state of the art
data- and pipeline-parallel training system.
We do not evaluate FTPipe\,\cite{eliad2021fine} (another pipeline-parallel training system),
as it is designed for the fine-tuning on commodity GPUs;
also its published prototype failed to run in different settings than their evaluated ones\,\cite{ftpipe_github}.

\subsubsection{Fine-Tuning Experiments}
We run the fine-tuning of RNN and BERT-24 on four V100 GPUs that are interconnected via NVLink.
The following four settings are mainly evaluated:
\textsf{a}) cross-layer model parallelism, 
\textsf{b}) GPipe, 
\textsf{c}) OOO-Pipe1 (GPipe with gradient fast-forwarding), and 
\textsf{d}) OOO-Pipe2 (OOO-Pipe1 with modulo allocation).
We also report the performance of PipeDream for a subset of the experiments.
We use the batch size of 1024 for RNN and 96 for BERT, which is commonly used for their fine-tuning\,\cite{shoeybi2019megatron,lan2019albert,yang2019xlnet,dai2020funnel}.
When applying modulo allocation for RNN, we assign $i$'th cell to GPU\textsubscript{$i$ mod $4$};
for BERT-24, we assign $i$'th encoder to GPU\textsubscript{$i$ mod $4$}.

\begin{figure}[t!]
\centering
\includegraphics[width=3.35in]{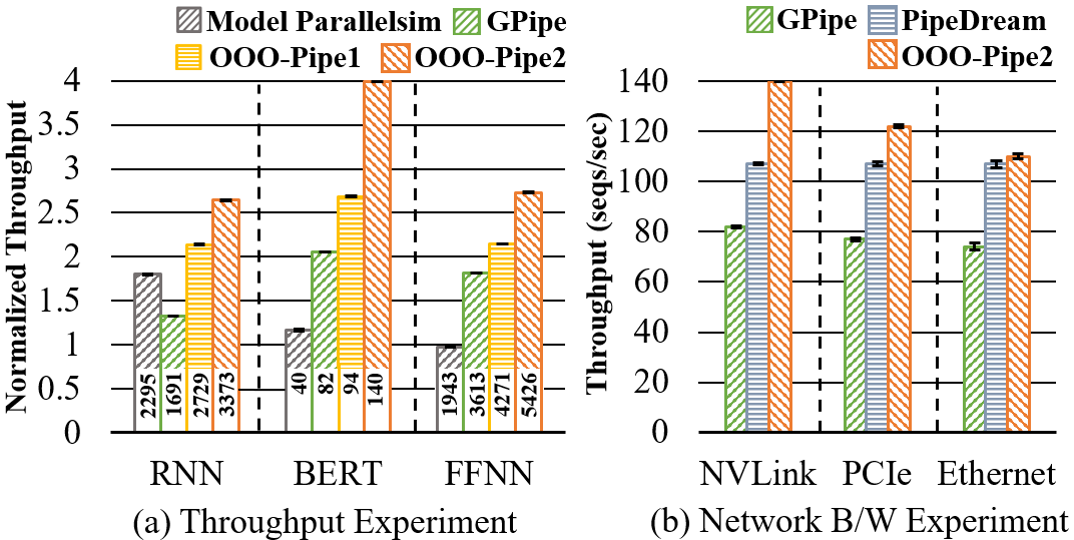}
\vspace{-2mm}
\caption{Throughput of NLP models for pipeline-parallel training on 4$\times$V100;\,(a) is the throughput for RNN, BERT, and FFNN normalized to that of single-GPU training, and (b) is BERT-24 with three interconnect networks. The numbers above x-axis are the actual throughput.\label{fig:pipeline_parallel}
}
\vspace{-4mm}
\end{figure}

Figure~\ref{fig:pipeline_parallel}\,(a) shows the results.
The x-axis shows the execution \textsf{a}--\textsf{d} for RNN and BERT, and y-axis is the throughput normalized by that of the single-GPU training.
The numbers on the x-axis are the actual throughput (sequences per second).
For the RNN model, OOO-Pipe2 runs 1.99$\times$ faster than the baseline (GPipe);
compared to (cross-layer) model parallelism, OOO-Pipe2 is 1.47$\times$ faster.
For BERT-24, OOO-Pipe1 runs 1.15$\times$ faster than the baseline
and together with modulo layer allocation OOO-Pipe2 is 1.59$\times$ faster.
Compared to the single-GPU training of BERT-24, we achieve 3.2$\times$ speedup with four GPUs.

\begin{figure*}[tbh]
\centering
\includegraphics[width=7.0in]{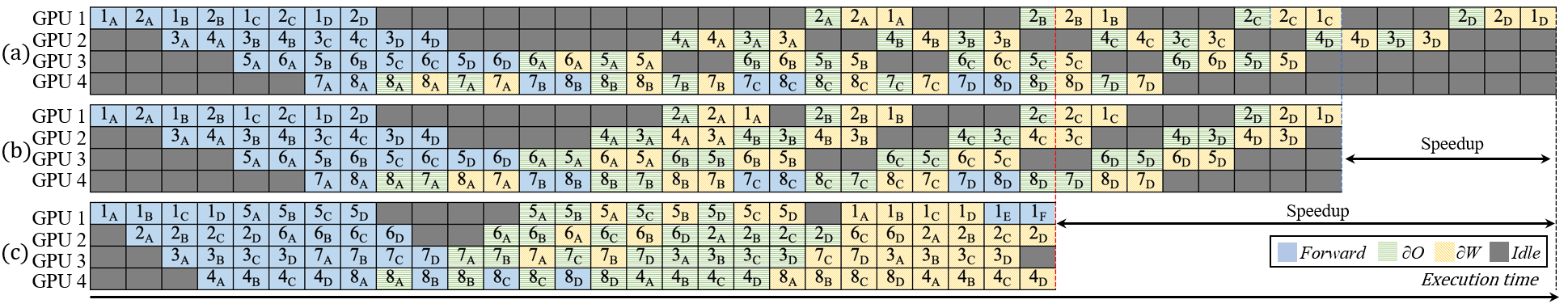}
\vspace{-2mm}
\caption{Execution timelines of pipeline-parallel training of 8-layer feed-forward neural network\,(FFNN). Execution\,(a) is GPipe, (b) is OOO-Pipe1 (gradient fast-forwarding), and (c) is OOO-Pipe2 (OOO-Pipe1 + modulo allocation). The numbers are the layer index and the subscript is
the micro-batch index.}
\label{fig:pipeline_timeline}
\vspace{-1mm}
\end{figure*}

To better understand the performance impact of the two optimizations, 
we also evaluate the training of a simple feed forward neural network (FFNN) with 16 fully-connected layers.
When applying the same set of optimizations, we obtain the experimental results
in Figure~\ref{fig:pipeline_parallel}\,(a) denoted by FFNN. 
With the two optimizations OOO-Pipe2 runs 1.5$\times$ faster than the baseline.
To study the effect of the two optimizations for FFNN, we illustrated the execution timelines with and without the optimizations.
For simplicity we assume that all the computations take the same amount of time and the inter-GPU communication latency is negligible.
We show the execution timelines for FFNN with 8 layers for the interest of space, but 
we report our analysis with the 16-layer FFNN.
Figure~\ref{fig:pipeline_timeline} shows the execution timelines (of 8-layer FFNN);
(a) is GPipe, (b) is OOO-Pipe1 (gradient fast-forwarding), and
(c) is OOO-Pipe2 (OOO-Pipe1 with modulo allocation).
The numbers denote the layer index of the computation and the subscript (A,B,C, and D) denotes the index of micro-batches.
If we examine the timelines for the 16-layer FFNN, 
gradient fast-forwarding gives 1.22$\times$ speedup over GPipe
and together with modulo allocation our execution is 1.62$\times$ faster than GPipe.
Compared to these, our experimental results yield reduced speedup (1.18$\times$ and 1.5$\times$) because of the inter-GPU communication overhead and non-uniform kernel execution times.

\noindent
{\bf Communication overhead.} We run another set of experiments to study the impact of the increased communication by modulo allocation.
We trained BERT-24 on four V100 GPUs with three different interconnect networks:
NVLink (50GB), PCIe 3.0 (16GB), and 10Gb Ethernet (1GB).
We measured the training throughput of GPipe, OOO-Pipe2, and PipeDream.
Figure~\ref{fig:pipeline_parallel}\,(b) shows the results.
In all three interconnect settings, OOO-Pipe2 substantially outperforms GPipe by 70\% in NVLink,
by 58\% in PCIe, and by 48\% in 10Gb Ethernet.
PipeDream applies {\it weight stashing} to train with 
multiple version of weight parameters, 
and thus the communication overhead of a small cluster is masked
by the computation with older or newer parameters.
In this four-GPU cluster, the communication overhead of OOO-Pipe2 seems 
relatively larger in low-bandwidth network.
However, in a typical cluster setting for training large NLP models,
where a small number of GPUs in a node are interconnected in high-bandwidth network
and the nodes are interconnected in low-bandwidth network,
the communication overhead for OOO-Pipe2 is not large, as we show in our pre-training evaluation.

\noindent
{\bf Memory overhead.} Gradient fast-forwarding incurs memory overhead for
storing the input tensors of the delayed computations.
When training the NLP models on four V100 GPUs, the maximum memory overhead incurred by
the fast-forwarding is 11\% over the baseline.
However, modulo allocation eliminates the memory overhead by immediately transferring
and discarding the computed outputs, 
thus using the same amount of memory as the baseline.

\subsubsection{Pre-Training Experiments}
We evaluate the pre-training of BERT-\{12,24,48\} and GPT-3 (Medium, 24 decoders)
on a larger cluster of 36$\times$V100 (Pub-B in Table~\ref{tab:clusters}).
We set the max sequence length for BERT be 128 and that for GPT-3 be 512 for the pre-training\,\cite{devlin2018bert,narayanan2020memory}. 
We used the vocabulary size of 30,522 for the BERT models and
50,257 for GPT-3, which is commonly used for these models\,\cite{lan2019albert,liu2019roberta,shoeybi2019megatron}.
As we are interested in the scalability of the training with increasing number of GPUs,
we perform both weak scaling and strong scaling experiments.
In the weak scaling experiments, we train with 8--32 GPUs and with BERT-12--BERT-48.
We evaluated three systems: GPipe, PipeDream, and OOO-Pipe2 (ours).
For this evaluation, we use the batch sizes of 512--1872 that give the maximum performance for each system.

\begin{figure}[h]
\centering
\includegraphics[width=3.3in]{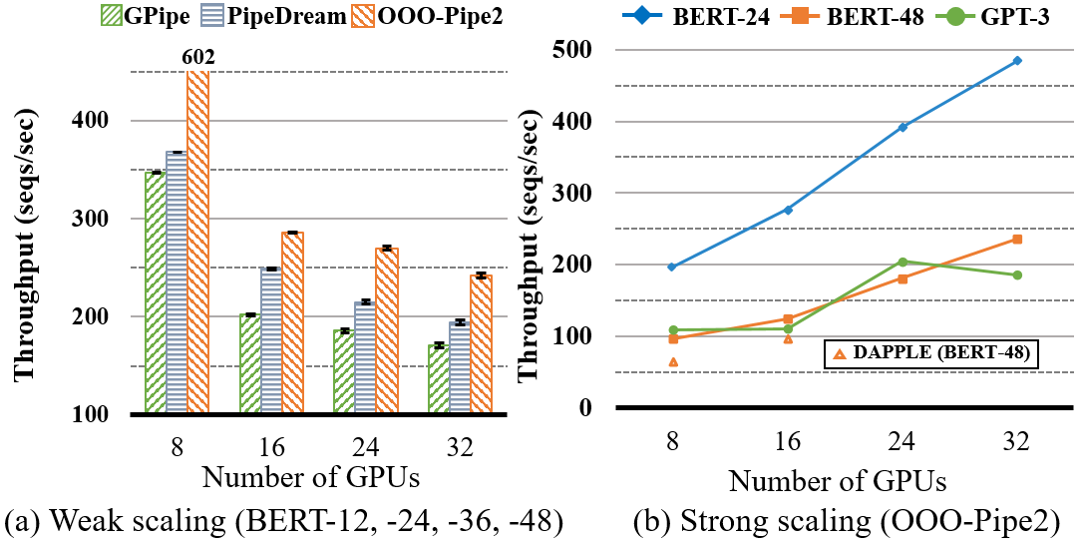}
\vspace{-1mm}
\caption{Testing the scalability of pipeline-parallel training. We used 8--32 GPUs for transformers; extra 4 GPUs are assigned for the word embedding layer in all settings.\label{fig:scaling}}
\vspace{-1mm}
\end{figure}

Figure~\ref{fig:scaling}(a) shows the results of the weak scaling experiments.
With eight GPUs, the transformers are all assigned to 
the GPUs on the same machine, which are interconnected via NVLink. In this case, OOO-Pipe2 is
1.73$\times$ faster than GPipe and 1.63$\times$ faster than PipeDream.
With 16 to 32 GPUs, we trained larger models (BERT-24 to BERT-48)
and the speedup by OOO-Pipe2 is 41-45\% over GPipe and 14-25\% over PipeDream.
When we increased the number of GPUs from 16 to 32, the performance gain of OOO-Pipe2 does not decrease but it either remains similar (compared to GPipe)
or increases (compared to PipeDream).
This demonstrates that the performance gain by modulo allocation 
outweighs the overhead of the increased communication for the evaluated cluster setting.
For training BERT-48 in PipeDream, we set the maximum number of parameter versions to be 32 as it gives 
the maximum training throughput. Training with high staleness level of 32 may negatively affect 
the learning efficiency and slow down the convergence\,\cite{ho2013more, dai2018toward}.

Now we describe the strong scaling experiments for OOO-Pipe2.
We trained BERT-\{24,48\} with 8--32 GPUs in the same way as the previous experiments.
For GPT-3, the size of its last embedding layer (including the intermediate tensors)
is large due to its large sequence length and vocabulary size, 
and thus we separately assign four GPUs to the layer, which is split in the output neuron dimension;
the four GPUs compute the last word embedding and the first embedding lookup operations.
Figure~\ref{fig:scaling}\,(b) shows the experimental results.
We observe that the performance of training BERT-24 and BERT-48 scales fairly well,
with the total number of GPUs for the training increasing from 8 to 32; 
with 4$\times$ number of GPUs, the throughput for the two BERT models is increased by 2.5$\times$.
For comparison we also evaluated Dapple for the pipeline-parallel training of BERT-48 with 8 and 16 GPUs; 
note that Dapple is evaluated with maximum 16 GPUs\,\cite{fan2021dapple} and their
published system does not support larger GPU clusters out-of-the-box\,\cite{dapple_github}. 
With 8 and 16 GPUs, OOO-Pipe2 is 1.47$\times$ and 1.29$\times$ faster than Dapple respectively. 
When we applied both data- and pipeline-parallel training to Dapple and OOO-Pipe2, the performance of the two systems similarly improved by 30--35\%.
However, finding the optimal hybrid-parallel training with ooo backprop is beyond the scope of this paper and we leave it as our future work.
For GPT-3, we used 12 to 36 GPUs with the extra 4 GPUs for the word token embedding layer. The performance for GPT-3 scales in a limited manner because it consists of 24 transformer decoders. 
That is, the 24 transformers cannot be evenly assigned to
16 or 32 GPUs, hence its scalability from 8 to 16 and
from 24 to 32 (transformer) GPUs is limited.

\section{Related Work}
\label{sec:related}

\noindent
{\bf Optimization of single-GPU training.}
Kernel fusion reduces the execution overhead of short-running kernels
by combining multiple consecutive kernels into a single one.
The technique is applied by optimizing compilers such as XLA\,\cite{leary2017xla, chen2018tvm, nvidia_fast_trans}.
In our evaluation, we show that our optimizations can be applied to XLA
to further speedup the executions of the fused kernels.
The technique of pre-compiling a group of kernel issues has been recently proposed
to reduce the kernel issue overhead\,\cite{kwon2020nimble,chen2015mxnet}.
Particularly, Nimble\,\cite{kwon2020nimble} also applies multiple GPU streams
for the neural network models that have parallel blocks such as Inception blocks\,\cite{szegedy2015going} to compute those blocks in parallel.
However, Nimble (and other systems supporting multiple GPU streams\,\cite{narayanan2018accelerating, yu2020salus}) do not concurrently
execute weight and output gradient computations in an out-of-order manner
as our proposed techniques.

\noindent
{\bf Optimization of data-parallel training.}
Hao et al. proposed wait-free backpropagation for data-parallel training\,\cite{poseidon},
which overlaps the parameter synchronization of a layer with the prior layer's
gradient computations.
More recently, the technique of prioritizing parameter communication has been proposed
to improve the performance by giving higher priority to the parameter communications
in critical path\,\cite{peng2019generic,jayarajan2019priority,byteps,hashemi2018tictac,li13pytorch}.
Out-of-order backprop further reorders the critical gradient computations 
as well as the critical synchronizations 
to more efficiently overlap the gradient computations with their communications.
Our scheduling algorithm based on out-of-order backprop
largely outperforms BytePS that 
prioritizes the parameter communications\,\cite{byteps}.

A number of studies have proposed relaxed schemes for 
parameter synchronization\,\cite{adpsgd2018, ssp2013, easgd2015}.
For example in AD-PSGD, 
a worker synchronizes its parameters with only one other
worker in each training iteration\,\cite{adpsgd2018}.
While these techniques reduce the synchronization overhead,
they are likely to incur accuracy loss in practice\,\cite{wongpanich2020rethinking,gupta2016model}.
Our optimizations do not change the semantics of the training 
and thus they can be safely and generally applied without incurring accuracy loss.

\noindent
{\bf Optimization of pipeline-parallel training.}
For the training of large neural networks that do not fit in a single GPU, pipeline-parallel training is proposed\,\cite{gpipe19, pipedream19, fan2021dapple, eliad2021fine}.
The proposed systems commonly apply the micro-batch technique that splits
a mini-batch into multiple micro-batches and pipelines the executions with those micro-batches.
Our two optimizations, i.e., gradient fast-forwarding and modulo layer allocation,
are applied on top of the micro-batch technique
to further improve the training performance, as we show in our evaluation.
PipeDream\,\cite{pipedream19} and FTPipe\,\cite{eliad2021fine} also support
asynchronous pipeline-parallel training,
which incurs staleness and thus change the semantics of the training
in a similar manner to asynchronous data-parallel training.

\section{Conclusion}
\label{sec:conclude}

This paper proposes out-of-order backprop, an effective
scheduling technique for neural network training.
By exploiting the dependencies of gradient computations,
out-of-order backprop reorders weight gradient computations to 
speed up neural network training.
We proposed optimized scheduling algorithms based on out-of-order backprop
for single-GPU, data-parallel, and pipeline-parallel training.
In single-GPU training, we schedule with multi-stream out-of-order computation
to mask the kernel execution overhead;
we also apply pre-compiled kernel issue to eliminate the kernel issue overhead.
In data-parallel training, we reorder the weight gradient computations 
to maximize the overlapping of computation and parameter synchronizations.
In pipeline-parallel training, we prioritize the execution of output gradient computations 
to reduce the pipeline stalls; we also apply modulo layer allocation, an optimized
layer allocation policy.
We implemented out-of-order backprop and all our scheduling optimizations in TensorFlow XLA and BytePS.
In our evaluation with twelve neural network models, 
our optimizations largely outperform the respective state of the art 
training techniques.
For single-GPU training, our optimizations outperform XLA by 1.03--1.58$\times$.
For data-parallel training, our technique is 1.1--1.27$\times$ faster than BytePS in large GPU clusters.
For pipeline-parallel training, our optimizations is 1.41--1.99$\times$ faster than GPipe and 
1.14--1.63$\times$ faster than PipeDream.

\bibliographystyle{plain}
\bibliography{paper}

\end{document}